\documentclass[runningheads]{llncs}
\usepackage[T1]{fontenc}
\usepackage{graphicx}
\usepackage{bbding}
\usepackage{amssymb}
\usepackage{multirow}
\usepackage{cite}
\usepackage{amsmath}
\usepackage{adjustbox}
\usepackage{hyperref}
\begin{document}
\title{Test-Time Intensity Consistency Adaptation for Shadow Detection}
\author{Leyi Zhu\inst{1}\orcidID{0009-0001-9783-7629} \and
Weihuang Liu\inst{1}\orcidID{0000-0002-9532-7633} \and 
Xinyi Chen\inst{2} \and
Zimeng Li\inst{3}\orcidID{0000-0003-2798-3134} \and
Xuhang Chen\inst{1,4}\orcidID{0000-0001-6000-3914} \and
Zhen Wang\inst{4}\orcidID{0000-0002-7262-2998} \and
Chi-Man Pun\inst{1 (}\Envelope\inst{)}\orcidID{0000-0003-1788-3746}
}
\authorrunning{L. Zhu et al.}
%\orcidID{0009-0001-9783-7629} \orcidID{1111-2222-3333-4444} \orcidID{2222--3333-4444-5555}
\institute{University of Macau, Macao, China \and
Southern University of Science and Technology, Shenzhen, China \and
Shenzhen Polytechnic University, Shenzhen, China \and
Huizhou University, Huizhou, China\\
\email{cmpun@umac.mo}}

% \institute{}

\newcommand{\lwh}[1]{\textcolor{red}{[lwh: #1]}}

\maketitle              % typeset the header of the contribution
\begin{abstract}
Shadow detection is crucial for accurate scene understanding in computer vision, yet it is challenged by the diverse appearances of shadows caused by variations in illumination, object geometry, and scene context. Deep learning models often struggle to generalize to real-world images due to the limited size and diversity of training datasets. To address this, we introduce TICA, a novel framework that leverages light-intensity information during test-time adaptation to enhance shadow detection accuracy.
TICA exploits the inherent inconsistencies in light intensity across shadow regions to guide the model toward a more consistent prediction. A basic encoder-decoder model is initially trained on a labeled dataset for shadow detection. Then, during the testing phase, the network is adjusted for each test sample by enforcing consistent intensity predictions between two augmented input image versions. This consistency training specifically targets both foreground and background intersection regions to identify shadow regions within images accurately for robust adaptation.
Extensive evaluations on the ISTD and SBU shadow detection datasets reveal that TICA significantly demonstrates that TICA outperforms existing state-of-the-art methods, achieving superior results in balanced error rate (BER).
\keywords{Shadow Detection \and Test-time Adaptation \and Consistency Training.}
\end{abstract}

\section{Introduction}

Shadow detection serves as an essential aspect of computer vision, which aims to accurately identify shadow regions within images. This identification is critical for distinguishing shadows from objects, preventing their misinterpretation, and ultimately enhancing the accuracy of subsequent image processing tasks such as semantic segementation~\cite{py1,py2,py3,chen15}, medical imaging~\cite{chen5,chen6,liu2020fine,chen17,chen18,chen20}, image enhancement~\cite{chen4,chen7,liu2023coordfill,zhang1,zhang4,chen16}, and scene understanding~\cite{liu2024dh,liu2023image,zhang3}.

Despite its importance, shadow detection faces significant challenges due to the complex interplay of factors influencing shadow formation. Variations in illumination conditions, object geometry, and scene context can lead to diverse shadow appearances~\cite{chen3,zhang2,zhang6}, making it difficult to develop robust and generalizable shadow detection models.

\begin{figure}[t]
\centering
\includegraphics[width=12cm]{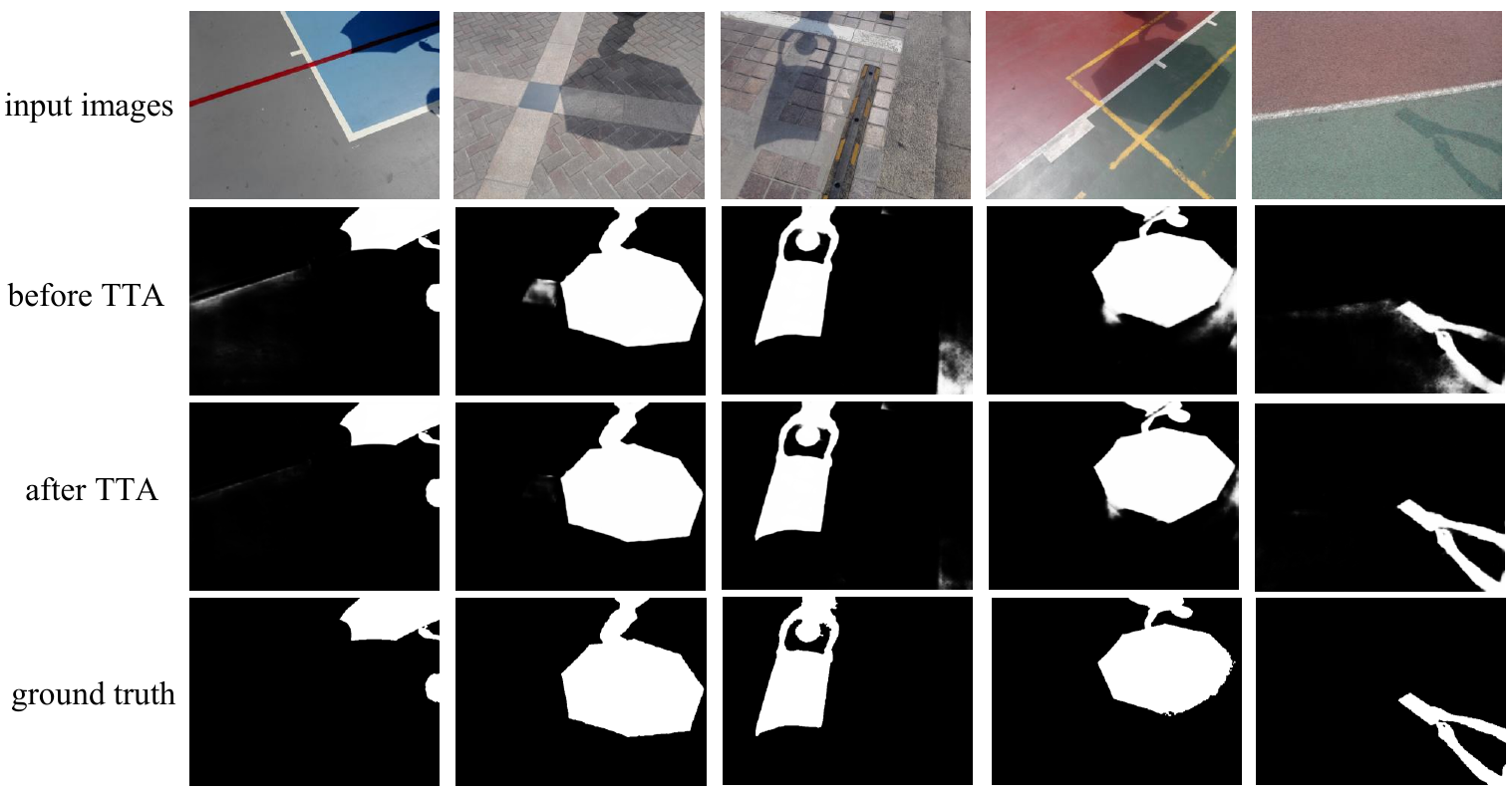}
\caption{\textbf{Light intensity inconsistency in shadow detection.} Row two displays the results from the pre-trained model. Without the TTA strategy, the model tends to identify dark areas as shadow regions. Following the application of our TICA strategy, the prediction masks show improved accuracy in detecting shadows.} 
\label{motivation}
\vspace{-1em}
\end{figure}

Traditional methods typically rely on analyzing real-world variations in light intensity, color, and texture to improve their performance~\cite{finlayson2009entropy,finlayson2005removal,guo2011single,guo2012paired}.
The advent of deep learning has changed this paradigm~\cite{zhang2022correction,jiang2020geometry,zhang7,yao2020speech,chen9,jiang2021deep,li2023cee}.
More recently, deep learning-based algorithms have shown promise by learning features directly from data~\cite{huang2011characterizes,lalonde2010detecting,zhu2010learning,liu2023explicit,chen1,liu2023explicit2,zhang8,li2022monocular}. However, the generalization capability of these approaches is often constrained by the lack of abundant, labeled training datasets and the inherent distribution shift between training and test data. This limitation hinders their performance on real-world images with unseen variations. What's more, shifts in light intensity have been shown to greatly impact the performance of shadow detection~\cite{zhu2021mitigating,chen2}, as shown in Fig.\ref{motivation}. Consequently, accurately detecting shadows demands a global comprehension of the image semantics~\cite{hu2018direction,zhang9,chen12,chen13,chen14}. 

To address these challenges, we draw inspiration from recent advances in test-time adaptation (TTA), which have demonstrated excellent performance in handling distribution shifts~\cite{sun2020test,wang2021tent,schneider2020improving,liu2024depth} by fine-tuning pre-trained models on each test sample using an unsupervised objective function. Building upon this, we propose a novel test-time intensity consistency adaptation (TICA) framework that leverages light-intensity information inherent in shadow images.

Our key insight is that inconsistencies in light intensity, often caused by varying relative positions of the light source and occluder, can be exploited to improve shadow detection. By enforcing consistency in light intensity between two random segmentations of the same input image during TTA, we make the model produce consistent responses across different shadow regions.

The proposed TICA utilizes a basic encoder-decoder network that undergoes a two-phase training process. In the initial stage, the model learns to predict shadow masks using standard supervised training. In the second stage, we apply data augmentation to each test sample and fine-tune the model by enforcing consistent responses within the intersection regions of these augmented views. This fine-tuned model is then used for improved shadow detection.
We evaluate TICA on two well-known public shadow detection datasets, ISTD~\cite{wang2018stacked} and SBU~\cite{vicente2016large}, and demonstrate its effectiveness in significantly improving shadow detection performance in comparison to existing state-of-the-art (SOTA) TTA solutions. 
% Notably, TICA achieves competitive results in terms of balance error rate (BER).

Our key contributions are highlighted as follows:
\begin{itemize}
\item[$\bullet$] We introduce TICA, the first TTA framework for shadow detection by leveraging light intensity consistency during testing. 
% To the best of our knowledge, this is the first work to demonstrate the effectiveness of incorporating intensity consistency constraints within a test-time adaptation approach for shadow detection.
\item[$\bullet$] We propose a novel consistency training strategy. This strategy employs a dual constraint on intensity consistency within the intersection area, considering both foreground and background regions for robust adaptation.
\item[$\bullet$] Our extensive experiments on two widely used public shadow detection datasets, demonstrating our method's effectiveness in comparison to SOTA TTA methods.
\end{itemize}

\section{Related Works}

This section reviews prior work relevant to our research on single-image shadow detection, focusing on two key areas: shadow detection methodologies and TTA techniques.

\subsection{Shadow Detection}

Image shadow detection (ISD) is a form of semantic segmentation aimed at distinguishing shadow from non-shadow pixels. Although general image object segmentation (IOS) methods are applicable to ISD, they typically perform poorly due to the specific characteristics of shadows. To address this, ISD has evolved into two primary strategies: traditional and deep learning-driven approaches.

Initially, ISD utilized physics-based models relying on illumination invariant assumptions \cite{finlayson2009entropy, finlayson2005removal}. These methods were effective under controlled lighting \cite{zheng2019distraction,nguyen2017shadow} but faltered with complex, real-world images. 
Later, data-driven statistical learning \cite{lalonde2010detecting} was adopted to improve shadow detection in diverse conditions, using hand-crafted features and various classifiers \cite{guo2011single, huang2011characterizes, lalonde2010detecting, zhu2010learning, vicente2015leave}. However, these methods were limited by the inadequacies of manually designed features in complex scenes.

% \subsubsection{Deep Learning-Based Methods}
The advent of deep learning~\cite{chen11,jiang2020geometry,zhang10,chen8,chen10,zhang5}, particularly convolutional neural networks (CNNs), has significantly enhanced ISD.
Early work by Khan \textit{et al.}~\cite{hameed2014automatic} employed a 7-layer ConvNet for the automatic extraction of features from superpixels, which were then refined using a Conditional Random Field model for the prediction of shadow detection. Vicente \textit{et al.}~\cite{vicente2016large} introduced a semantically-aware patch-level ConvNet that incorporated image-level semantic information during training.
Recognizing the importance of contextual information for accurate shadow detection, researchers began exploring methods to incorporate such information into deep learning models. Nguyen \textit{et al.}~\cite{nguyen2017shadow} launched scGAN, leveraging a tunable sensitivity parameter in the generator to effectively produce shadow masks. The ability of the scGAN generator to process the full image enabled it to capture global context and structure. Hu \textit{et al.}~\cite{hu2018direction} proposed a novel direction-aware attention mechanism within a spatial recurrent neural network to analyze image context for improved shadow detection.
More recently, researchers have investigated the integration of lighting information as an additional prior for enhancing the performance of shadow detectors. Le \textit{et al.}~\cite{le2018a+} presented a framework that combines a shadow detection network with a shadow attenuation network designed to produce adversarial training samples, leading to improved detection performance. Zhu \textit{et al.}~\cite{zhu2021mitigating} designed a feature decomposition and reweighting mechanism to reduce intensity inconsistency in shadowed regions. Chen \textit{et al.}~\cite{chen2020multi} introduced a teacher-student framework~\cite{tarvainen2017mean} that leveraged unlabeled data and explored the simultaneous learning of multiple shadow-related information cues.

\subsection{Test-Time Adaptation}

TTA is a paradigm focused on adapting a pre-trained model to unlabeled examples in the testing phase to enhance its generalization capabilities. Building upon advancements in refining pre-trained models using test samples for image classification~\cite{schneider2020improving,wang2021tent,niu2022efficient}, our research explores the application of TTA techniques to the shadow detection problem within the context of image object segmentation tasks.
Unlike test-time training (TTT)~\cite{sun2020test}, which requires defining a proxy task and designing a proxy model architecture for both training and testing, TTA methods can directly adapt the pre-trained model using target task data without such additional efforts.

Various TTA techniques have been proposed in the literature. Schneider \textit{et al.}~\cite{schneider2020improving} introduced a simple yet effective approach that replaces batch normalization statistics computed from the training set with those computed from the test data. Wang \textit{et al.}~\cite{wang2021tent} proposed TENT, an entropy-based adaptation technique that minimizes entropy to fine-tune the normalization layers of the model during testing. Niu \textit{et al.}~\cite{niu2022efficient} incorporated a Fisher regularizer to constrain significant changes in important model parameters during adaptation. They further extended this work by proposing a method to filter out samples with large gradients, encouraging model weights to converge to a minimum~\cite{niu2023towards}. In the realm of energy-based adaptation, Yuan \textit{et al.}~\cite{yuan2024tea} applied an energy function to implicitly capture the underlying data distribution, facilitating alignment with the test data distribution.
As far as we are aware, the application of TTA to image object segmentation for shadow detection remains an unexplored area, which our research aims to address.

\section{Method}

\begin{figure}[t]
\includegraphics[width=\textwidth]{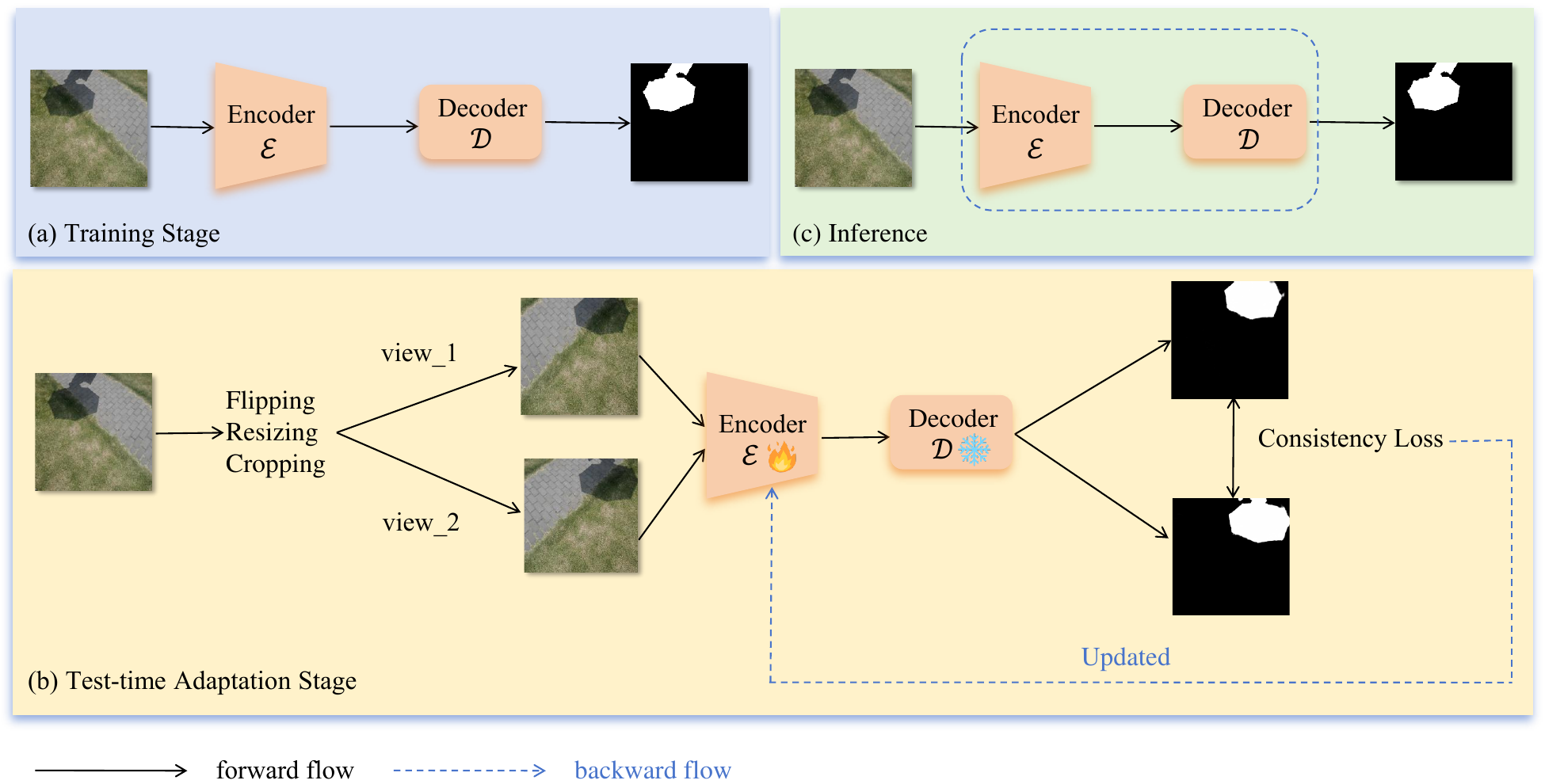}
\caption{\textbf{Overview of the proposed TICA.} By leveraging light consistency training, the TICA framework enhances the model's capabilities in shadow detection. Initially, the model is trained with a publicly accessible shadow detection dataset. We then apply random data augmentation techniques—horizontal flipping, resizing, and cropping—to the test set. This facilitates model refinement by enforcing consistent intensity predictions between the two augmented images. The consistency loss is backpropagated to update the encoder.} 
\label{overview}
\vspace{-1em}
\end{figure}

This section details our TICA framework for shadow detection. Figure \ref{overview} illustrates the overview of the pipeline.
Divided into two stages, our approach starts with developing a standard encoder-decoder network aimed at shadow detection in the training stage. During testing stage, the model is updated by enforcing intensity consistency to make a better prediction. 
% In the first stage, we train TICA on a public shadow detection dataset to generate shadow masks using the decoder $\mathcal{D}_{ta}$. 
% Subsequently, during the test-time adaptation (TTA) stage, the model enforces intensity consistency between two augmented samples derived from the same input image. This process involves optimizing an objective function and updating the parameters of both the encoder $\mathcal{E}_{te}$ and decoder $\mathcal{D}_{te}$, thereby enhancing the model's ability to discern shadow regions. This iterative refinement ultimately yields a refined shadow prediction mask. It is important to note that the parameters of $\mathcal{E}_{te}$ and $\mathcal{D}_{te}$ are initialized from the final epoch parameters of $\mathcal{E}_{ta}$ and $\mathcal{D}_{ta}$, respectively.

% We begin by elaborating on the training and test-time adaptation stages in Sections \ref{sec:3.1} and \ref{sec:3.2}, respectively. Section \ref{sec:3.3} then delves into the intensity consistency training strategy.

\newcommand{\encoder}{\mathcal{E}}
\newcommand{\decoder}{\mathcal{D}}
\subsection{Training Stage}\label{sec:3.1}

% \subsubsection{Objective Function}
Fig. \ref{overview}(a) illustrates the training stage framework. We employ a common encoder-decoder network for shadow detection. 
Our initial step is extracting multi-scale features from the input image via the encoder~$\encoder$ and then feeding these features into the decoder~$\decoder$ to predict the shadow mask.
Following~\cite{zheng2019distraction,zhu2021mitigating}, we utilize the balanced binary cross-entropy (BBCE) loss \cite{hu2018direction} during training.
% To address the imbalanced distribution of shadow and non-shadow regions in natural images, we utilize the balanced binary cross-entropy (BBCE) loss \cite{hu2018direction} during training.
% Let $\mathnormal{y}$ denote the ground truth value of a pixel, where 1 represents a shadow region and 0 represents a non-shadow region. Let $\widetilde{\mathnormal{y}} \in [0, 1]$ represent the predicted probability of a pixel belonging to a shadow region. The BBCE loss function $\mathcal{L}_{bbce}$ is then given by:
%$\widetilde{\mathnormal{Y}}$

We denote the ground truth as $\mathnormal{Y}$, and $\mathnormal{y}$ represent the predicted shadow mask, the BBCE loss $\mathcal{L}_{bbce}$ is then given by:
\begin{equation}
\mathcal{L}_{bbce} = -\sum_i \left( \frac{N_p}{N_p+N_n} \mathnormal{Y_i} \log(\mathnormal{y_i}) + \frac{N_n}{N_p+N_n} (1-\mathnormal{Y_i}) \log(1-\mathnormal{y_i}) \right),
\label{bbce}
\end{equation}
where \textit{i} indexes across every pixel in the image, with $N_p$ and $N_n$ representing the count of shadow pixels and non-shadow pixels, respectively. The BBCE loss encourages the model to assign equal importance to detecting both shadow and non-shadow regions, mitigating bias towards the majority class. Note that $\mathcal{L}_{bbce}$ is computed at the pixel level and then summed over all pixels.

\subsection{TTA Stage}\label{sec:3.2}

% \subsubsection{Intensity Consistency TTA}
% During test-time adaptation, we employ mini-batches of four input images. 
As depicted in Fig. \ref{overview}(b), the encoder $\encoder$ is updated by the proposed TICA, which jointly optimizes the consistency between predictions from two different augmented views of each input image. Recognizing that accurate shadow detection relies on both foreground and background information, our TTA strategy leverages predictions from both perspectives. 

Specifically, for each test image, we first generate two different views $i^1$ and $i^2$ using different random transformations (flipping, resizing, and cropping). Then, we predict the corresponding shadow mask $y^1$ and $y^2$.
Next, we calculate the foreground consistency (FC) loss via:
\begin{equation}
\mathcal{L}_{fg} = \mathcal{L}_{KL}(y^1_i, y^2_i), \forall i \in (y^1 \cap y^2) 
\label{fg_loss}
\end{equation}
where $\mathcal{L}_{KL}$ denotes the Kullback-Leibler (KL) divergence\cite{bu2018estimation}. We binarize the predicted mask using a threshold of 0.5 when obtaining the intersection region.
We also calculate the background consistency (BC) loss as follows:
\begin{equation}
\mathcal{L}_{bg} = \mathcal{L}_{KL}(\overline{y^1_i}, \overline{y^2_i}), \forall i \in (\overline{y^1} \cap \overline{y^2}),
\label{bg_loss}
\end{equation}
where $\overline{y^1} = 1 - y^1$ and $\overline{y^2} = 1 - y^2$.

Overall, TICA optimizes the model by considering a dual constraint on both foreground and background intensity consistency:
\begin{equation}
\mathcal{L}_{tica} = \lambda_{fg} \mathcal{L}_{fg} + \lambda_{bg} \mathcal{L}_{bg},
\label{total_loss}
\end{equation}
where $\lambda_{fg}$ and $\lambda_{bg}$ are hyper-parameters balancing the foreground loss $\mathcal{L}_{fg}$ and the background loss $\mathcal{L}_{bg}$, respectively. In our experiments, we observed that simultaneously updating the encoder and decoder did not yield better model performance and negatively impacted the efficiency of the TTA stage. Consequently, we decided to update only the encoder $\encoder$ during the TTA phase.

% During TTA, we only update the encoder $\encoder$.

By incorporating information from both foreground and background intersections, we capture global image context, enabling more effective fine-tuning of the model's prediction response. This intensity consistency adaptation strategy encourages the model to produce robust and consistent shadow detection masks across different image regions.

\section{Experiments}
\subsection{Datasets and Evaluation Metrics}
\subsubsection{Datasets} Our evaluation of TICA is conducted on two publicly available datasets, ISTD and SBU. The ISTD dataset comprises 1,330 images for the training set and 540 images for the test set, while the SBU dataset includes 4,089 images for the training set and 638 images for the test set, respectively. For the evaluation on the ISTD test set, we initially train a baseline model with the ISTD training set and then fine-tune the model on the ISTD test set by TICA. The evaluation on SBU dataset also follows this same flow.

\subsubsection{Evaluation Metrics} 
The BER serves as the performance metric for quantitative comparisons with other state-of-the-art methods in our analysis. The formulation of BER is as follows:
\begin{equation}
    BER = 1- \frac{1}{2}(\frac{TP}{TP + FN}+\frac{TN}{TN+ FP}),
\label{ber}
\end{equation}
where $TP$, $FP$, $TN$, and $FN$ represent the counts of true positives, false positives, true negatives, and false negatives, respectively. We calculate the BER for the shadow and non-shadow regions separately, averaging the results for the final BER.

\subsection{Implementation Details}
We select ResNet-50 \cite{he2016deep}, Swin-Tiny \cite{liu2021swin}, and HRNet-18 \cite{sun2019deep} as the image encoders, while the segmentation decoder is built using the SegFormer’s lightweight decoder \cite{xie2021segformer}.

\subsubsection{Training stage details.} The encoder $\mathcal{E}$ and decoder $\mathcal{D}$ are trained using the AdamW optimizer~\cite{kingma2014adam}. For the ISTD dataset, training is conducted for 50 epochs, whereas the training epoch is 20 for the SBU dataset. The initial learning rate is configured to $2 \times 10^{-4}$ for the ISTD dataset and $5 \times 10^{-5}$ for the SBU dataset. During training, the initial learning rate is adjusted using a cosine decay schedule. Input images are scaled to $400 \times 400$ pixels. Data augmentation techniques, including random horizontal flipping, resizing, and cropping, are applied to the training set.

\subsubsection{TTA stage details.} 
During the TTA stage, only the encoder $\mathcal{E}$ is updated using the Adam optimizer. The learning rate during this adaptation phase is configured to $1 \times 10^{-5}$ for the ISTD dataset and $5 \times 10^{-7}$ for the SBU dataset. Input images are adjusted to a resolution of $400 \times 400$ pixels before being processed by the adapted models. All TTA methods are trained for 5 epochs. Data augmentation, including random horizontal flipping, resizing, and cropping, is applied to the test samples. Experimentally, the hyperparameters $\lambda_{fg}$ and $\lambda_{bg}$ are set to 0.5 and 1, respectively, for the ISTD dataset, and to 1 and 0.05, respectively, for the SBU dataset.

% %baseline: flip+resize+crop; tta:flip+resize+crop
% \begin{table}[h]
% \centering
% \caption{Quantitative comparison of the intensity consistency on different backbone models across two public datasets.}
% \label{ablation_of_method}
% \begin{tabular}{|l|l|l|l|l|}
% \hline
% Backbone & Method & ISTD~\cite{wang2018stacked} & SBU~\cite{vicente2016large}\\
% \hline
% \multirow{4}{*}{ResNet-50~\cite{he2016deep}} & - & 2.4498 & 6.0651\\
% & w/ foreground consistency & 1.9345 &  \textbf{4.6982}\\
% & w/ background consistency & 1.6493  & 4.8320 \\
% & w/ both & \textbf{1.6409} & 4.7214\\ %resnet epoch is last 3 but the best is 3
% \hline
% \multirow{4}{*}{Swin-Tiny~\cite{liu2021swin}} & - & 1.7569 & 4.4381 \\
% & w/ foreground consistency & 1.9592 & \textbf{3.9329}\\
% & w/ background consistency & \textbf{1.5791} & 3.9795\\
% & w/ both & 1.5855 & 3.9367\\ %%swin epoch is last 3 but the best is 2
% \hline
% \multirow{4}{*}{HRNet-18~\cite{sun2019deep}} & - & 1.4721 & 5.1439\\
% & w/ foreground consistency & 1.4324 & \textbf{4.3691} \\
% & w/ background consistency & \textbf{1.1655} & 4.4747 \\
% & w/ both & 1.1677 &  4.3841 \\ %hrnet epoch is last 3 but the best is 2
% \hline
% \end{tabular}
% \end{table}

%Compare the BER of the intensity consistency on different backbone models across two public datasets.

\begin{table}[ht]
\centering
\caption{Ablation study using variable intensity consistency strategy on different backbone models across two public datasets.}
\label{ablation_of_method}
\begin{tabular}{c|cccc|cccc|cccc}
\hline
\multirow{2}{*}{} & \multicolumn{4}{c|}{ResNet-50 as Backbone} & \multicolumn{4}{c|}{Swin-Tiny as Backbone} & \multicolumn{4}{c}{HRNet-18 as Backbone} \\ \cline{2-13} 
                  & -     & FC  & BC & both          & -     & FC  & BC & both          & -    & FC & BC & both          \\ \hline
ISTD              & 2.45  & 1.93     & 1.65    & \textbf{1.64} & 1.76  & 1.96     & 1.58    & \textbf{1.59} & 1.47 & 1.43    & 1.17    & \textbf{1.17} \\
SBU               & 6.07  & 4.70     & 4.83    & \textbf{4.72} & 4.44  & 3.93     & 3.98    & \textbf{3.94} & 5.14 & 4.37    & 4.47    & \textbf{4.38} \\ \hline
\end{tabular}
\end{table}

\subsection{Analysis and Ablation Studies}\label{sec:4.3}

\subsubsection{Effects of TTA Strategy.}
We begin by analyzing the proposed TICA strategy. To verify the effectiveness of this approach, we evaluate our complete method against the following variants:

(\uppercase{\romannumeral1}) a vanilla baseline without the intensity consistency loss, where we utilize the BBCE loss as defined in Eq. \eqref{bbce} during the training phase to train the model; (\uppercase{\romannumeral2}) a variant that refines the model during TTA using only the foreground intersection region loss in Eq. \eqref{fg_loss}; (\uppercase{\romannumeral3}) a variant that uses only the background intersection region loss in Eq. \eqref{bg_loss} for TTA; (\uppercase{\romannumeral4}) a variant that leverages both the foreground and background intersection region losses in Eq. \eqref{total_loss} for TTA.

Table \ref{ablation_of_method} presents the results of the performance evaluation. Incorporating the intensity consistency loss to refine the model during TTA results in superior performance compared to the baseline across three different backbones. Balancing the foreground and background intersection losses facilitates the use of consistency loss for model updates, thereby enhancing shadow detection. The results underscore the importance of light intensity information for improving shadow detection performance.

\begin{figure}[t]
\centering
\includegraphics[width=12cm]{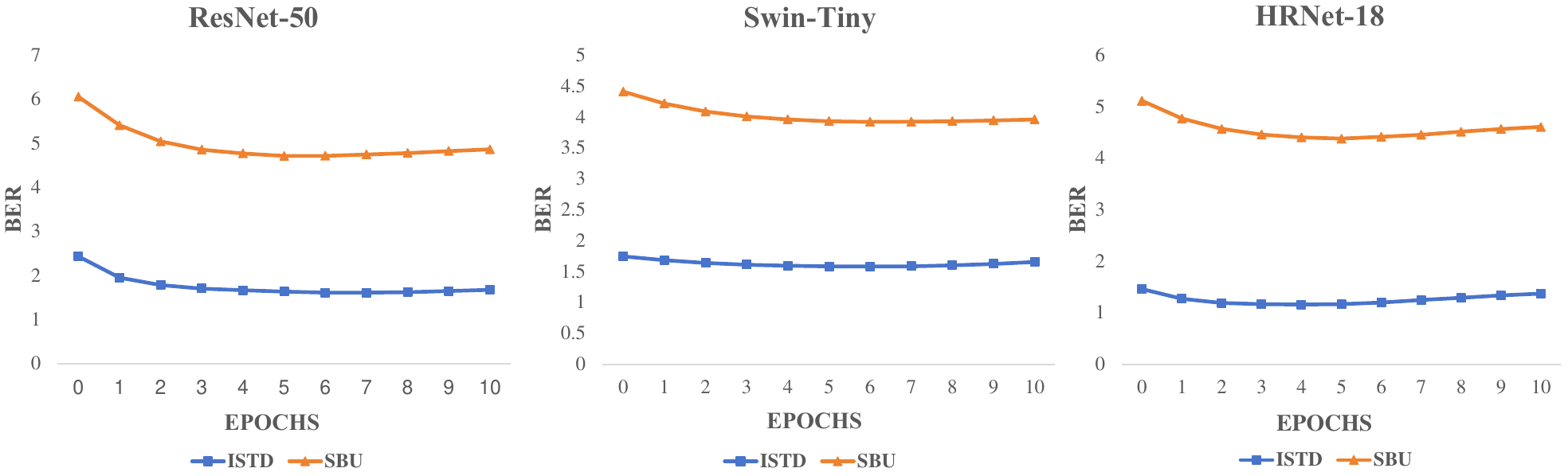}
\caption{\textbf{The impact of the proposed TICA fluctuates with the number of epochs.} Our TICA strategy is evaluated on the ISTD and SBU datasets across three backbone architectures: ResNet-50, Swin-Tiny, and HRNet-18. 
%We can observe that the BER rate starts to drop initially and becomes steady after the fifth epoch.
% The proposed TICA yields enhanced prediction performance by the fifth epoch.
} 
\label{training_epochs}
\vspace{-1em}
\end{figure}

\begin{figure}[t]
\centering
\includegraphics[width=12cm]{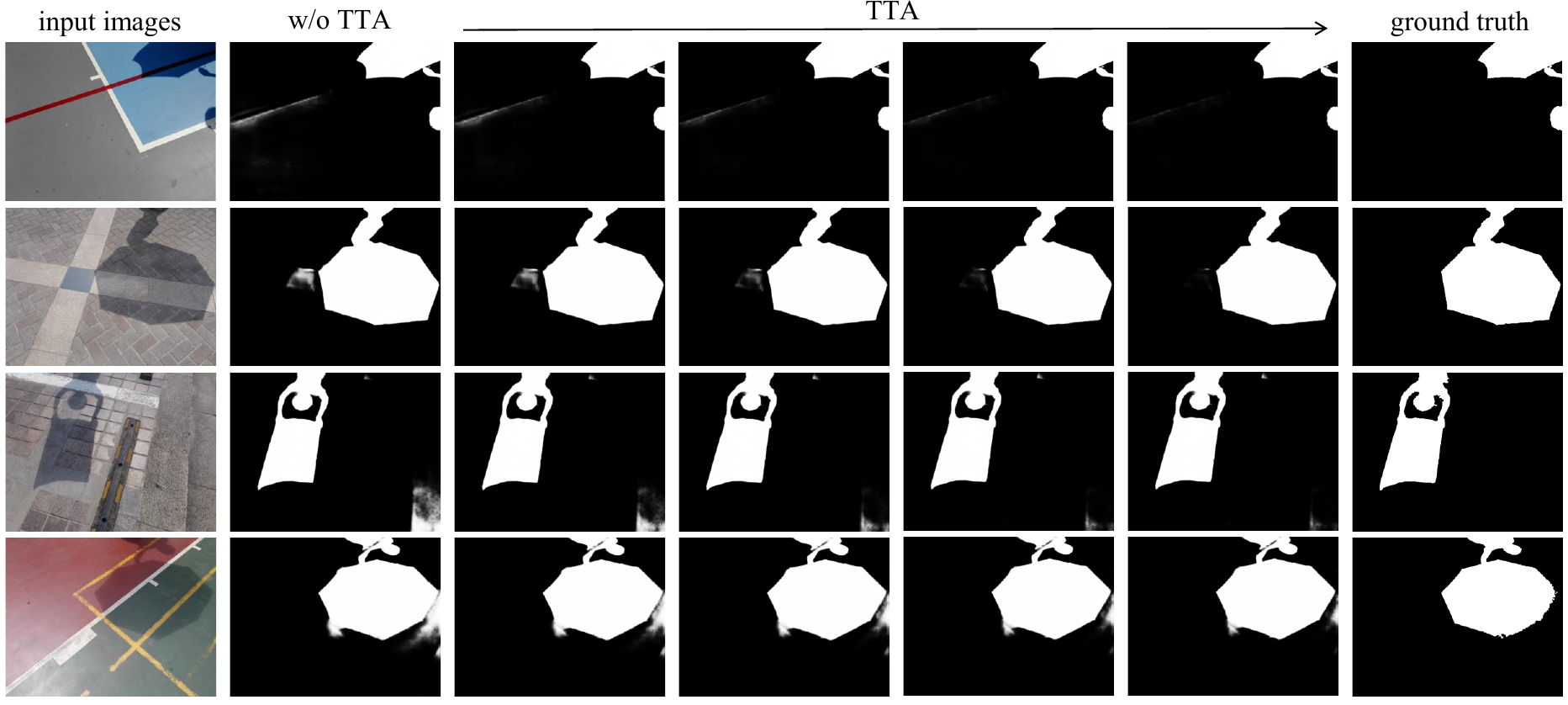}
\caption{\textbf{Qualitative comparison of the fine-tuning results and ground truth for TICA over five epochs.} We observe that the original prediction masks from the pre-trained model are coarse, with results improving as the TTA epochs increase (within 5 epochs).} 
\label{tta}
\vspace{-1em}
\end{figure}

\subsubsection{Training Epochs.}
One significant challenge in employing TTA is the increased computational time required. To investigate the impact of training epochs on TTA, we set the maximum epochs at 10 and monitored the model's performance. The results are illustrated in Fig. \ref{training_epochs}. With an increase in the number of training epochs, the model's performance first enhances but then starts to deteriorate. Moreover, the optimal training epochs fluctuate depending on the dataset. Consequently, selecting a suitable number of epochs, such as 5, can yield satisfactory performance on multiple datasets.
Fig. \ref{tta} presents the detection results of our method. The pre-trained model from the training stage demonstrates limited capability in detecting shadows. However, after applying TICA, our method significantly improves the model's capacity for effective shadow detection.

\subsection{Comparison with the SOTA TTA Techniques}
% We compare with two post-processing methods and three previous test-time adaptation methods in Table \ref{compare_with_ttt}, including 
% CRF~\cite{lafferty2001conditional}, 
% BS~\cite{barron2016fast}, 
We present a comparison of our method with three previous TTA techniques in Table \ref{compare_with_ttt}, including TENT~\cite{wang2021tent}, ETA~\cite{niu2022efficient}, and BN~\cite{schneider2020improving}. BN adaptation~\cite{barron2016fast} updates the mean and variance for each batch of test samples. TENT minimizes the entropy of test samples during test time. ETA selects low-entropy samples to optimize the model parameters. Our TICA method ensures consistent intensity in both foreground and background by leveraging global information to adapt and refine shadow detection.

\begin{table}[h]
\caption{Compare the BER with the SOTA TTA methods for shadow detection.}
\label{compare_with_ttt}
\adjustbox{width=\linewidth}{
\begin{tabular}{l|ccccc|ccccc|ccccc}
\hline
\multirow{2}{*}{}         & \multicolumn{5}{c|}{ResNet-50 as Backbone}      & \multicolumn{5}{c|}{Swin-Tiny as Backbone}      & \multicolumn{5}{c}{HRNet-18 as Backbone}        \\ \cline{2-16} 
                          & -    & TENT & ETA  & BN   & Ours          & -    & TENT & ETA  & BN   & Ours          & -    & TENT & ETA  & BN   & Ours          \\ \hline
\multicolumn{1}{c|}{ISTD} & 2.45 & 1.97 & 1.82 & 1.79 & \textbf{1.64} & 1.76 & 1.97 & 1.75 & 1.73 & \textbf{1.59} & 1.47 & 1.29 & 1.20 & 1.20 & \textbf{1.17} \\
\multicolumn{1}{c|}{SBU}  & 6.07 & 5.46 & 5.18 & 5.41 & \textbf{4.72} & 4.44 & 4.02 & 4.04 & 4.32 & \textbf{3.94} & 5.14 & 5.40 & 4.93 & 5.55 & \textbf{4.38} \\ \hline
\end{tabular}}
\vspace{-1em}
\end{table}

To compare the performance of different methods, we apply these methods to the baseline model. For TTA, the loss is evaluated for a batch of test samples during each iteration, and the adaptation requires multiple training epochs. All TTA methods are configured with a batch size of 4 and 5 epochs. As shown in Table \ref{compare_with_ttt}, our TICA method achieves the best BER scores among SOTA TTA methods on two datasets. On the ISTD dataset, we reduce the BER scores by 8.36\%, 8.43\%, and 2.44\% compared to the second best-performing methods using ResNet-50, Swin-Tiny, and HRNet-18, respectively. For the SBU dataset, we achieve improvements in BER scores of 8.87\%, 1.97\%, and 11.04\% against the second-best performer using ResNet-50, Swin-Tiny, and HRNet-18, respectively.

Existing TTA approaches have primarily focused on addressing data distribution discrepancies in corrupted datasets and natural distribution shifts. However, light intensity proves to be a critical variable in the process of detecting shadows in images~\cite{zhu2021mitigating}. We utilize a dual constraint on intensity consistency within the foreground and background to enforce the pre-trained model's adaptation to different areas' intensity. The results verify that our test-time intensity consistency strategy significantly improves shadow detection performance.

\subsection{Comparison with the SOTA Shadow Detectors}
We conduct a comparative evaluation with recent shadow detection methods. For a fair comparison, we directly use the results of the compared methods as presented in their respective published papers.
In Table \ref{compare_with_sota_shadow_detectors}, we present a summary quantitative results of our method and the SOTA methods on two public datasets, with performance reported in terms of BER. It is noteworthy that our TICA approach does not require additional training data, whereas some techniques incorporate supplementary training data for better performance on the SBU dataset. The results demonstrate that TICA attains optimal performance on the ISTD dataset with the HRNet-18 backbone. Compared to the second-best technique, FDRNet, TICA reduces the BER score by 24.5\% on the ISTD dataset.

%flip+resize+crop
\begin{table}[t]
\centering
\caption{Comparing our method with the SOTA shadow detectors.}
\label{compare_with_sota_shadow_detectors}
{
\begin{tabular}{|c|c|c|c|}
\hline
 Method & Backbone & {ISTD} & {SBU} \\
\hline
% Method & Backbone & BER $\downarrow$ & BER $\downarrow$ \\
\hline
DSC~\cite{hu2018direction}           & VGG16~\cite{vgg16} & 3.42      & 5.59   \\
DSD~\cite{zheng2019distraction} & ResNeXt-101~\cite{xie2017aggregated} & 2.17 & 3.45 \\
MTMT~\cite{chen2020multi} & ResNeXt-101~\cite{xie2017aggregated}  & 1.72 & 3.15 \\
FDRNet~\cite{zhu2021mitigating} & EfficientNet-B3~\cite{tan2019efficientnet} & 1.55  & \textbf{3.04}\\ %others ttt record the epoch is 10 in sbu but the tica best epoch is 5(rerun, i final record the epoch 5 for other methods)
% SDCM~\cite{zhu2022single} &  EfficientNet-B3~\cite{tan2019efficientnet} & 1.44 & \bf2.94 \\
FCSDNet~\cite{valanarasu2023fine} &  ResNeXt101~\cite{xie2017aggregated} & 1.69 & 3.15 \\
\hline
\multirow{3}{*}{Ours}  & ResNet-50~\cite{he2016deep}  & 1.64  & 4.72  \\
& Swin-Tiny~\cite{liu2021swin}  & 1.59 &  3.94\\
& HRNet-18~\cite{sun2019deep}  & \textbf{1.17}  & 4.38 \\
\hline
\end{tabular}}
\end{table}

\begin{figure}[t]
\includegraphics[width=\textwidth]{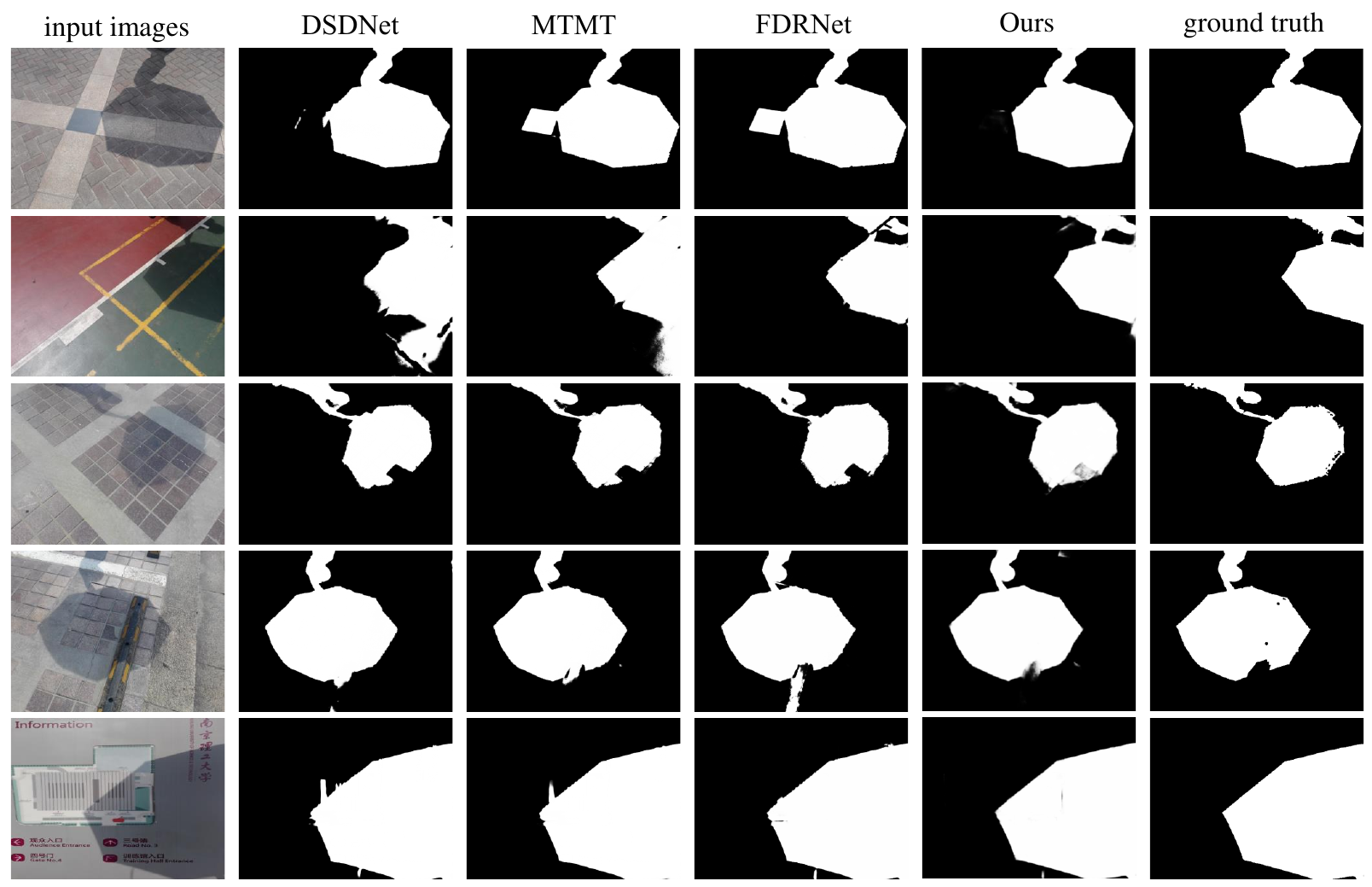}
\caption{\textbf{Visual comparison of other SOTA Shadow Detectors and our method (TICA) against ground truth's shadow mask.} It is evident that our method can detect shadow regions more accurately. Our results in the fifth column exhibit greater consistency with the ground truth in the sixth column than the other methods shown in the second to fourth columns on the ISTD dataset.} 
\label{tica_visual}
\vspace{-1em}
\end{figure} %figure should mark the citation!!!

% \subsubsection{Visual Comparison.}
Furthermore, we demonstrate visual results to conduct a qualitative comparison of our method against other existing approaches. Our method has demonstrated advantages in distinguishing between shadow regions and dark regions. Regions with dark colors present a significant challenge for other techniques to accurately recognize the actual shadow areas. Existing methods often struggle to effectively differentiate between shadows and dark regions. In comparison, our TICA relies on both foreground and background information to observe the holistic image semantics, resulting in enhanced shadow detection capabilities. As shown in Fig. \ref{tica_visual}, the visual comparisons are depicted.

\section{Conclusion}
This paper proposes TICA, a novel framework designed to enhance shadow detection accuracy by leveraging light-intensity information during the TTA process. TICA effectively addresses the challenge of diverse shadow appearances caused by variations in illumination, object geometry, and scene context, which often hinder the generalization capability of deep learning models trained on limited datasets.
The effectiveness of TICA stems from its ability to exploit inconsistencies in light intensity across shadow regions. By enforcing consistent intensity predictions between augmented versions of the input image during TTA, the model is guided toward a more accurate and robust representation of shadow boundaries.
Our extensive evaluations on the ISTD and SBU shadow detection datasets demonstrate TICA's superior performance compared to existing SOTA methods, achieving significant improvements. Notably, TICA's success highlights the importance of incorporating light intensity consistency as a valuable constraint during TTA for shadow detection.
Future research directions include exploring the integration of other shadow-related cues, such as texture and color, into the TICA framework for further performance enhancement. Additionally, investigating the applicability of TICA to other image segmentation tasks where intensity consistency plays a crucial role presents a promising avenue for future work.

\section*{Acknowledgment}
This work was supported by the University of Macau under Grants MYRG2022-00190-FST and MYRG-GRG2023-00131-FST, in part by the Science and Technology Development Fund, Macau SAR, under Grants 0141/2023/RIA2 and 0193/2023/RIA3.

\bibliographystyle{splncs04}
\bibliography{1496}




% LNCS Homepage, \url{http://www.springer.com/lncs}, last accessed 2023/10/25
% \end{thebibliography}
\end{document}